\documentclass[letterpaper]{article} 
\usepackage[]{aaai24}  
\usepackage{times}  
\usepackage{helvet}  
\usepackage{courier}  
\usepackage[hyphens]{url}  
\usepackage{graphicx} 
\usepackage{natbib}  
\usepackage{caption} 
\usepackage{algorithm}
\usepackage{algorithmic}

\usepackage[utf8]{inputenc} 
\usepackage[T1]{fontenc}    
\usepackage{url}            
\usepackage{booktabs}       
\usepackage{amsfonts}       
\usepackage{nicefrac}       
\usepackage{microtype}      
\usepackage{amsthm}
\usepackage{amsmath}
\usepackage{graphicx}
\usepackage{amssymb}
\usepackage{diagbox}
\usepackage{multirow}
\usepackage{pifont} 
\usepackage{bm}
\usepackage{subcaption}
\usepackage{color,xcolor}
\usepackage[font=small,labelfont=bf]{caption}  
\usepackage{makecell}
\usepackage{tabulary}
\definecolor{demphcolor}{RGB}{144,144,144}

\definecolor{mygray}{gray}{0.4}
\usepackage{colortbl}
\usepackage{enumitem}
\usepackage{multirow}
\usepackage{threeparttable}
\usepackage{multirow}
\usepackage[export]{adjustbox}

\newcommand{\tabincell}[2]{\begin{tabular}{@{}#1@{}}#2\end{tabular}}

\usepackage{newfloat}
\usepackage{listings}
\DeclareCaptionStyle{ruled}{labelfont=normalfont,labelsep=colon,strut=off} 
\floatstyle{ruled}
\newfloat{listing}{tb}{lst}{}
\floatname{listing}{Listing}
\title{Attend Who is Weak: Enhancing \\
Graph Condensation via Cross-Free Adversarial Training 
}
\author{
    Xinglin Li \textsuperscript{\rm 1}, Kun Wang \textsuperscript{\rm 2,*}, Hanhui Deng \textsuperscript{\rm 3}, Yuxuan Liang \textsuperscript{\rm 4},  Di Wu \textsuperscript{\rm 5,*} \\
}
\affiliations{
    \textsuperscript{\rm 1} lixinglin@hnu.edu.cn \;\;\;
    \textsuperscript{\rm 2} wk520529@mail.ustc.edu.cn \;\;\;
     \textsuperscript{\rm 3} denghanhui@hnu.edu.cn \\
     \textsuperscript{\rm 4} yuxliang@outlook.com \;\;\;
     \textsuperscript{\rm 5} dwu@hnu.edu.cn  \;\;\;
    \textsuperscript{\rm *} Corresponding authors  

}

\usepackage{bibentry}

\begin{document}

\maketitle

\begin{abstract}
\vspace{-0.3em}
In this paper, we study the \textit{graph condensation} problem by compressing the large, complex graph into a concise, synthetic representation that preserves the most essential and discriminative information of structure and features. We seminally propose the concept of \textbf{Shock Absorber} (a type of perturbation) that enhances the robustness and stability of the original graphs against changes in an adversarial training fashion. Concretely, (I) we forcibly match the gradients between pre-selected graph neural networks (GNNs) trained on a synthetic, simplified graph and the original training graph at regularly spaced intervals. (II) Before each update synthetic graph point, a Shock Absorber serves as a gradient attacker to maximize the distance between the synthetic dataset and the original graph by selectively perturbing the parts that are underrepresented or insufficiently informative. We iteratively repeat the above two processes (I and II) in an adversarial training fashion to maintain the highly-informative context without losing correlation with the original dataset. More importantly, \textit{our shock absorber and the synthesized graph parallelly share the backward process in a free training manner. Compared to the original adversarial training, it introduces almost no additional time overhead.}

We validate our framework across 8 datasets (3 graph and 5 node classification datasets) and achieve prominent results: for example, on Cora, Citeseer and Ogbn-Arxiv, we can gain nearly $1.13\%\sim 5.03\%$ improvements compare with SOTA models. Moreover, our algorithm adds only about 0.2\% to 2.2\% additional time overhead over Flicker, Citeseer and Ogbn-Arxiv. Compared to the general adversarial training, our approach improves time efficiency by nearly 4-fold. \textbf{The code is available in the supplementary material.}


\end{abstract}

\section{Introduction}

Graphs serve as a ubiquitous representation for a diverse range of real-world data, encompassing domains such as social networks \cite{guo2020deep, he2017neural, wang2019neural}, chemical molecules \cite{wang2022molecular, yang2023molerec, 10.1093/bib/bbac560}, transportation systems \cite{hong2020heteta, jin2023spatio}, and recommender systems \cite{wu2022graph}, among many others. As the tailor-made designs, Graph neural networks (GNNs) \cite{kipf2016semi, hamilton2017inductive, velivckovic2017graph, dwivedi2020benchmarking} have become a prevalent solution for machine learning tasks on graph-structured data, and have exhibited outstanding performance across a broad spectrum of graph-related applications \cite{zhou2020graph, ji2021survey, yue2020graph, gupta2021graph}.

However, real-world scenarios often involve large-scale graphs with millions of nodes and edges \cite{hu2020open, li2020deepergcn}, which presents significant computational overheads when training GNNs \cite{xu2018powerful, you2020l2, duan2022comprehensive}. Worse still, fine-tuning hyperparameters and identifying suitable training schemes for self-supervised models can be both expensive and resource-intensive, particularly for large-scale graphs with dense connections. To this end, when a GNN is making a prediction, one naturally raises a question: \textit{is it possible to effectively simplify or reduce the graph to not only accelerate graph algorithms, including GNNs, but also aid in storage, visualization, and retrieval for associated graph data analysis tasks \cite{jin2021graph, jin2022condensing, toader2019graphless, zhang2021graph}?}

To address this inefficiency, existing approaches typically fall into two research lines -- \textbf{graph sampling} and \textbf{graph distillation}. Within the first class, many endeavors \cite{chen2018fastgcn, eden2018provable, chen2021unified, sui2022inductive, gao2019graph, lee2019self} have investigated the use of custom-built sampling approaches to reduce the computational footprint of GNNs (including some pruning methods). These methods aim to identify the discriminative edges or nodes to enhance training or inference efficiency. Nevertheless, sampling or pruning graph nodes or edges may cause massive information loss, resulting in performance collapse \cite{wu2022structural, wangsearching}. To this end, many studies focus on the graph distillation research line. In contrast to simplifying the graph structure or nodes, the second research line targets condensing the large original graph into a small, synthetic, and highly informative graph. The objective is to train GNNs on the condensed graph, such that their performance is comparable to those trained on the original graph \cite{ying2018hierarchical, roy2021structure, ranjan2020asap, jin2021graph, jin2022condensing}. It is worth emphasizing that there are fewer prior studies on pruning or compressing GNNs \cite{bahri2021binary, wang2021bi, tailor2020degree}, which can bring the salient benefits for power-efficient graph representation learning. However, they cannot be extracted and modeled in data-level optimization, which goes out of the scope of our work.

\begin{figure*}[htbp]
\centering
\includegraphics[width=0.99\textwidth]{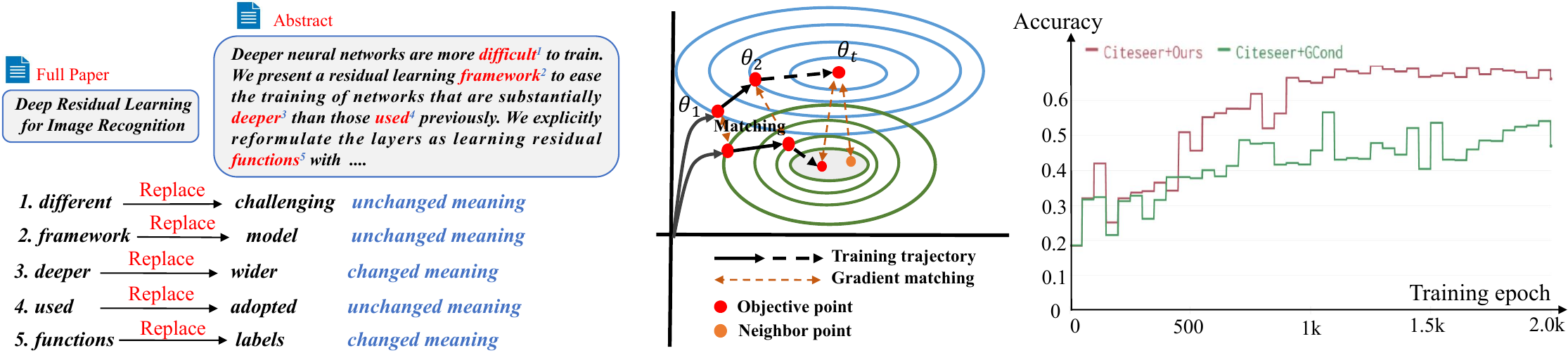}
\vspace{-0.5em}
\caption{\textbf{\textit{Left:}} The motivation of our proposal. \textbf{\textit{Middle:}} Traditional gradient matching can find a synthetic dataset with the closest parameters to the original dataset, near the red point, and the synthesized data (yellow point) to be close to the original dataset. However, the original gradient matching algorithm cannot effectively explore this neighborhood, resulting in a lack of resilience and robustness in the training process. \textbf{\textit{Right:}} The training example of our proposal on Citeseer dataset (compared with SOTA model GCond).}
\label{fig:intro}
\vspace{-1.5em}
\end{figure*}

Generally, graph distillation draws inspiration from data distillation techniques \cite{wang2018dataset, nguyen2021dataset, liu2022dataset, bohdal2020flexible, nguyen2020dataset} and aims to ensure consistency between raw and synthetic datasets by constraining the soft labels across both sets. Recently, some trajectory matching algorithms show great prominence in image \cite{kim2022dataset, lee2022dataset, jiang2022delving} and graph realms \cite{jin2021graph, jin2022condensing}. Concretely, these frameworks adopt parameter or gradient matching scheme w.r.t the condensed set and raw data during the training process. Though promising, enforcing gradient matching results in an inelastic compression process, as shown in Fig. \ref{fig:intro} (\textbf{\textit{Left}}). When we summarize a paper into an abstract, we can easily see that replacing some words does not change the meaning of abstract. The appearance of these synonyms in the synthetic data set will not change the meaning, but the trajectory matching expects that the synthetic set is \textbf{\textbf{completely}} consistent with the original set under each step of the training process. 

\textbf{Research gap.} Given a more intuitive instance (Fig.~\ref{fig:intro} (\textbf{\textit{Middle}})), graph trajectory matching algorithms enforce gradient consistency and provide a rigid parameter space that limits the flexibility during the training. In fact, these methods may not explore the impact of certain parameters that are similar in the vicinity of the optimal matching point.

This paper targets at overcoming this tricky hurdle and explores a more robust graph condensation framework. We propose a \textit{graph robust condensation algorithm, (\textbf{GroC})}, a principled adversarial training (bi-level optimization) framework that explores the neighborhood space of the parameters that have the greatest impact on the original matching process. To achieve this, we painstakingly design a \textbf{Shock Absorber} operator to attach perturbations of adversarial training in specific positioned location of the synthetic dataset. To highlight, the optimization process of our GroC is: (i) \textbf{robust,} as the training compression process is more stable and robust, which can better find a compressed subset (see the example in Fig.~\ref{fig:intro} \textbf{\textit{Right}}); (ii) \textbf{one-stop-shop,} since it is completely free of human labor of trial-and-error on perturbations and locations choices. (iii) \textbf{time-efficient,} through free training  algorithm, we parallelize the entire process of optimizing adversarial perturbations and synthesizing datasets, ensuring that virtually no additional time overhead is introduced.

\textbf{Contributions.} Our contributions can be summarized as follows: (1) We propose a robust adversarial training-based graph condensation framework called GroC for more effectively learning the robust representation space of the synthetic data, by attaching perturbations on the synthetic graph during the gradient matching process. (2) Shock Absorber operator can not only help our model achieve prominent performances, but also serve as a general operator for other compression frameworks. (3) Building on our insights, we train our framework on graph/node classification tasks. Stunningly, our model can yield SOTA results on various graph benchmarks, \textit{e.g.,} for example, on Cora, Citeseer and Ogbn-Arxiv, we can gain nearly $1.13\%\sim 5.03\%$ improvements compare with SOTA models under a smaller variance. Moreover, our algorithm adds only about 0.2\% to 2.2\% additional time overhead over Flicker, Citeseer and Ogbn-Arxiv. These results empirically demonstrate the effectiveness and robustness of our proposal.


\vspace{-0.4em}
\section{Preliminaries \& Related Work}
\vspace{-0.4em}



\textbf{Graph Neural Networks (GNNs).} Graph neural networks (GNNs) \cite{kipf2016semi, hamilton2017inductive, velivckovic2017graph, dwivedi2020benchmarking, wu2020comprehensive} are capable of processing variable-sized, permutation-invariant graphs. They learn low-dimensional representations through an iterative process that involves transferring and aggregating the representations from topological neighbors. Although GNNs have shown promising results, they face significant inefficiencies when scaling up to large or dense graphs. Towards this end, several research streams have focused on addressing this issue, such as graph sampling and graph distillation.




\noindent \textbf{Graph Sampling \& Distillation.} Graph sampling reduces the computational burden of GNNs by selectively sampling sub-graphs or applying pruning methods \cite{chen2018fastgcn, eden2018provable, chen2021unified, sui2022inductive, gao2019graph, lee2019self}. However, the aggressive sampling strategy may lead to a significant loss of information, potentially reducing the representation ability of the sampled subset. To address this, graph distillation research line \cite{ying2018hierarchical, roy2021structure, ranjan2020asap} draws inspiration from \textbf{d}ataset \textbf{d}istillation (DD), which aims to distill (compress) the knowledge embedded in raw data into synthetic data, ensuring that models trained on this synthetic data maintain performance \cite{wang2018dataset, zhao2023dataset, cazenavette2022dataset, nguyen2021dataset}. Remarkably, \cite{jin2021graph, jin2022condensing} take the first step to propose optimizing both nodes and edges in the graph by employing training gradient matching, which under the spotlight of our research.


\begin{figure*}[t]
\centering
\includegraphics[width=0.99\textwidth]{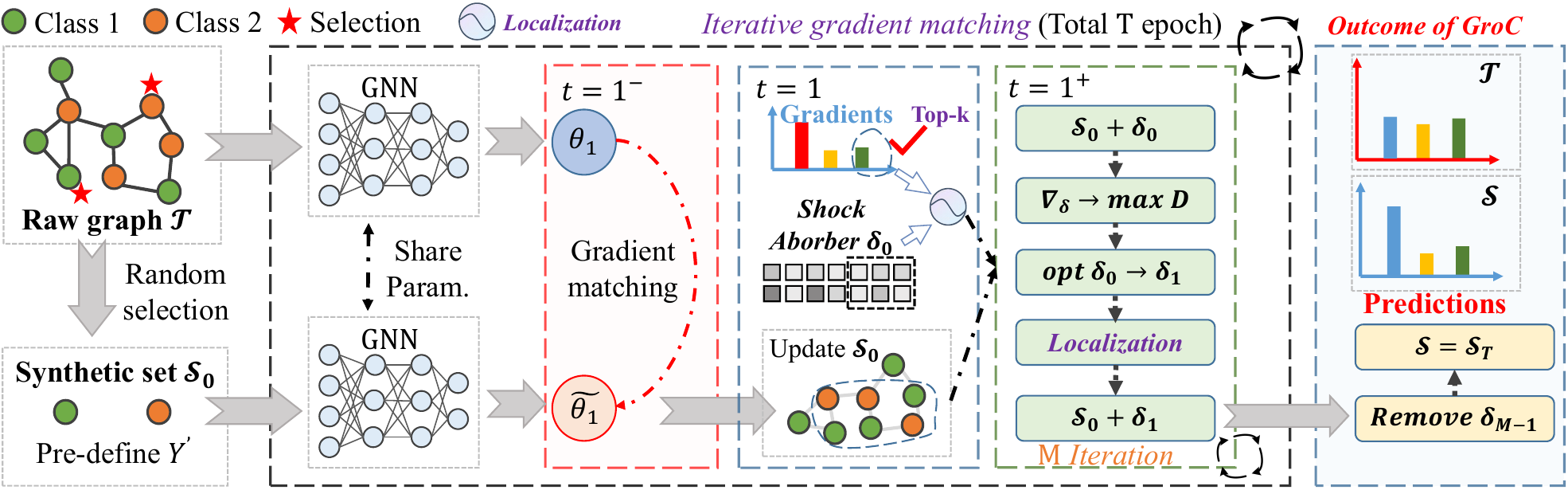}
\vspace{-0.5em}
\caption{Illustration of our GroC framework.}
\label{fig:method}
\vspace{-1.5em}
\end{figure*}

\noindent \textbf{Adversarial training \& Robustness.} Adversarial training was introduced as a defense mechanism against adversarial attacks, where a model is trained not only on the clean data but also on adversarial samples generated during training \cite{goodfellow2014explaining, moosavi2016deepfool}. They demonstrated that adversarial training can make deep neural networks more robust. Build upon these observations, many subsequent studies pay attention to design different adversarial examples \cite{papernot2016transferability, madry2017towards, chen2018fast, tramer2017ensemble}. In our work, we introduce existing methods to the inelasticity of synthetic data. Following the Projected Gradient Descent (PGD) \cite{madry2017towards}, performs iterative gradient descent with backtracking to generate adversarial perturbations.

\vspace{-0.4em}
\section{Methodology}
\vspace{-0.4em}

As shown in Fig. \ref{fig:method}, in this section, we devote to explaining how our GroC framework deepens and enhances the robustness in graph condensation task. Going beyond this, we take a comprehensive look at our key operator \textbf{Shock Absorber}. Finally, we introduce the free training for the efficient implementation GroC. In the following parts, we will delineate our model components by introducing Fig. \ref{fig:method} from left to right. For ease of understanding, we summarize all notations and depict our algorithm in Appendix, which can be checked in supplemental materials.


\vspace{-0.4em}
\subsection{Graph Condensation via Gradient Matching}
\vspace{-0.3em}

In this work, we first review the process of graph condensation and start from the original graph ${\rm{{\cal T}}}=(\rm{A}, \rm{X}, \rm{Y})$, where ${\rm{A}} \in \mathbb{R}^{N \times N}$ is the adjacency matrix, $N$ is the number of nodes, ${\rm{X}} \in \mathbb{R}^{N \times d}$ is the $d$-dimensional node feature attributes. Similar to traditional graph condensation task \cite{jin2021graph, jin2022condensing}, we note the label of nodes as ${\rm{Y}}={\left\{ {0,1, \ldots , C - 1} \right\}^N}$ denotes the node labels over $C$ classes. Our work target to train a synthetic graph $\rm{{\cal S}}=(\rm{A}^\prime,\rm{X}^\prime,{\rm{Y}^\prime})$ with adjacency matrix ${\rm{A'}} \in \mathbb{R}^{N' \times N'}$ and feature attributes ${\rm{X'}} \in \mathbb{R}^{N' \times D}$. We designate GroC, to obtain a synthetic graph with $N' \ll N$, which satisfies that a general GNN trained on $\cal S$ can reach commensurate accuracy on large graph $\cal T$.


\textbf{Gradient Matching as the Objective.} As our goal is to learn highly informative synthetic graphs, one prominent approach is to enable GNNs trained on synthetic graphs to mimic the training trajectory on the original large data. To achieve this goal, dataset condensation \cite{zhao2021dataset,zhao2020dataset,jin2021graph,jin2022condensing} introduces a gradient matching scheme. More specifically, \textit{it tries to minimize the discrepancy between the gradients of the model with respect to the parameters, as computed on a large-real data ${\rm{{\cal T}}}$ and a small-synthetic data ${\rm{{\cal S}}}$, at each training step.} Therefore, the parameters of the model trained on synthetic data will closely resemble those trained on real data at every training steps. We first formalize the problem as:

\vspace{-0.7em}
\begin{equation}\footnotesize
 \begin{aligned}
    & \mathop {\min }\limits_{\cal S} {\cal L}\left( {{\rm{GN}}{{\rm{N}}_{{\theta _{\cal S}}}}\left( {{\rm{A}},{\rm{X}}} \right),{\rm{Y}}} \right)\;\; \\ 
    & \textbf{s.t.} \;\; {\theta _{\cal S}} = {\rm{arg\;}}\mathop {{\rm{min}}}\limits_\theta  \;{\cal L} ( {{\rm{GN}}{{\rm{N}}_\theta } ( {{\rm{A'}},{\rm{X'}}} ),{\rm{Y'}}} )\;\;\;
 \end{aligned}
\end{equation}
\vspace{-0.7em}

\noindent where ${\rm{GN}}{{\rm{N}}_\theta }$ stands for the GNN initizalization with $\theta$, $\cal L$ represents the loss function. $\theta_{\cal S}$ denotes that model parameters trained on  $\cal S$. The labels of the synthetic graph are pre-defined. First, we generate a specific number of labels, ensuring an equal number of labels per class. Then, we randomly select the corresponding nodes from the original dataset to serve as the initial features for each class in the synthetic dataset. Following the \cite{jin2021graph}, we employ multiple initializations to mitigate the risk of over-fitting. For ease of understanding, we describe the gradient matching procedure based on a single initialization in the following part, unless otherwise specified.  

\vspace{-0.7em}
\begin{equation} \label{match}
\footnotesize
\begin{aligned}
& \mathop {\min }\limits_{\rm{{\cal S}}} \sum\limits_{t = 0}^T {D\left( {{\nabla _\theta } \ell _t^{\rm{{\cal S}}} ({f_{{\theta _t}}}({{\rm{{\cal S}}}_t}),{{\rm{Y}}^\prime }),{\nabla _\theta } \ell _t^{\rm{{\cal T}}} ({f_{{\theta _t}}}({\rm{{\cal T}}}),{\rm{Y}})} \right)} \\
& \quad \textbf{s.t.} \quad {\theta _{t + 1}} = \rm{opt}({\theta _t},{\rm{{\cal S}_t}})
 \end{aligned}
\end{equation} 
\vspace{-0.7em}

\noindent In Eq.~\ref{match}, $D (\cdot,\cdot)$ is a distance function, $f_{\theta _t}$ denotes the GNN model parameterized with ${\theta}$ at time point $t$. ${{\rm{{\cal S}}}_t}$ is the synthetic data of the $t$-th iteration of optimization. $T$ represents the total number of steps in the training process, and $\rm{opt}(\cdot,\cdot)$ is the optimization operator used for updating the parameters $\theta$.  This equation represents a bi-level problem, where we learn the synthetic graphs ${\rm{{\cal S}}}$ in the outer optimization loop, and update the model parameters ${\theta _t}$ in the inner optimization loop. $\ell _t^{\rm{{\cal S}}}$ and $\ell _t^{\rm{{\cal T}}}$  are the negative log-likelihood loss of the synthetic and original datasets, respectively. We conduct gradient matching process at different time points and our parameter update forms of two datasets can be written as:

\vspace{-0.7em}
\begin{equation} \footnotesize
\begin{aligned}
    & \theta _{t + 1}^{\rm{{\cal S}}} = {\rm{op}}{{\rm{t}}_\theta }\left( {\ell _t^{\rm{{\cal S}}} \left( {{\rm{GN}}{{\rm{N}}_{\theta _t^{\rm{{\cal S}}}}}\left( {{{\rm{A}}^\prime },{{\rm{X}}^\prime }} \right),{{\rm{Y}}^\prime }} \right)} \right) \\
    & \theta _{t + 1}^{\rm{{\cal T}}} = {\rm{op}}{{\rm{t}}_\theta }\left( {\ell _t^{\rm{{\cal T}}} \left( {{\rm{GN}}{{\rm{N}}_{\theta _t^{\rm{{\cal T}}}}}\left( {{\rm{A}},{\rm{X}}} \right),{\rm{Y}}} \right)} \right)
    \label{eq.inner_para}
\end{aligned}
\end{equation}

\vspace{-0.5em}

\noindent Here we use backpropagation to update the model parameters $\theta _{t + 1}^{\rm{{\cal S}}}$ and $\theta _{t + 1}^{\rm{{\cal T}}}$, and we proceed to consider to match the gradient distance of two graph set. Similar to \cite{jin2021graph}, we define distance $D$ as the sum of the distances $dis$ at each layer. For a specific layer, given two GNN model gradients ${G^{\rm{{\cal S}}}} \in {\mathbb{R}^{{d_1} \times {d_2}}} $ and ${G^{\rm{{\cal T}}}} \in {\mathbb{R}^{{d_1} \times {d_2}}}$, the distance $dis(\cdot, \cdot)$ used for condensation is defined as follows.

\vspace{-0.7em}
\begin{equation} \label{distance}
\footnotesize
\rm{dis}({G^{\rm{{\cal S}}}},{G^{\rm{{\cal T}}}}) = \sum\limits_{i = 1}^{d2} {( {1 - \frac{{G_i^{\rm{{\cal S}}} \cdot G_i^{\rm{{\cal T}}}}}{{\left\| {G_i^{\rm{{\cal S}}}} \right\|\left\| {G_i^{\rm{{\cal T}}}} \right\|}}} )}
 \end{equation} 
 \vspace{-0.5em}

\noindent In Eq. \ref{distance}, ${G^{\rm{{\cal S}}}}$ and ${G^{\rm{{\cal T}}}}$ represent the $i$-th column vectors of the gradient matrices. By employing these formulations, we can efficiently achieve gradient matching strategy.

However, traditional methods are notably unstable, and an excessive emphasis on gradient consistency can lead to synthesized graphs often lacking the desired generalization capability. \textbf{A promising approach is to introduce adversarial training, enabling the model to explore a wider latent space during its training process.} To this end, we introduce adversarial training into the graph condensation research line for the first time, setting the context for our investigation.

In the following parts, for the sake of convenience, we will draw on the concept of left and right limits from mathematical notation. We denote the superscript $t^{+}$ for the right limit and the superscript $t^{-}$ for the left limit \footnote{Assuming that $t^{+}$ is the right limit for $t$, for any $\varsigma > 0$ satisfies that $t^{+}-t > \varsigma $, and we can draw the similar conclusion in left limit $t^{-}$: $t-t^{-} > \varsigma $.}. Fig.~\ref{fig:method} showcases our GroC algorithms, in the initial stage, we do not update the GNN parameters. Instead, we optimize the trainable synthesized graph (outer loop). We refer to this particular time as the left limit of t,~\textit{e.g.,} $t^{-}$, in which we optimally update the synthetic dataset ${{\cal S}_{t - 1}} \to {{\cal S}_t}$ using the gradient computed by the GNN:

\vspace{-0.5em}
\begin{equation} \label{updateX}
\footnotesize    
 {\rm{X}^\prime } \leftarrow {\rm{X}^\prime } -  \eta _1{\nabla _{{\rm{X}^\prime }}}{D^\prime } \;\; {\rm{\textbf{if}}} \;\; t\;\% \left( {{{\rm{{\omega}}}_1} + {{\rm{{\omega}}}_2}} \right) < {{\rm{{\omega}}}_1}
\end{equation}
\vspace{-0.5em}


\noindent In Eq.~\ref{updateX}, here $D^\prime$ is the updated distance of the two datasets, we leverage the gradient distances to propagate optimized features of the synthetic dataset in the backpropagation fashion. However, frequently matching gradients may lead the whole optimization process notoriously time-consuming. As a trade-off, we match the gradient at regular intervals of $\omega_1 + \omega_2$  periodically. Concretely, in every $\omega_1 + \omega_2$ epoch, we match gradient $\omega_1$ times for optimizing feature attributes and the next $\omega_2$ epoch we only update adjacency matrix $\rm{A}^\prime$:

\vspace{-0.6em}
\begin{equation}\footnotesize
\begin{aligned}
        & {g_\phi }  \leftarrow {g_\phi }  - {\eta _2}{\nabla _\phi }{D^\prime } \;\;\Rightarrow \;\;  {\rm{A}^\prime } = {g_\phi }\left( {{\rm{X}^\prime }} \right) \;\; \\
        &  \text{with} \;\; {\rm{A}}_{ij}^{'} = {\sigma}( {( {{\rm{ML}}{{\rm{P}}_\phi }( {[ {{\rm{X}}_i^{'};{\rm{X}}_j^{'}} ]} ) + {\rm{ML}}{{\rm{P}}_\phi }( {[ {{\rm{X}}_j^{'};{\rm{X}}_i^{'}} ]} )} )/2} )
\end{aligned}
\end{equation}
\vspace{-0.5em}

\noindent Here $g_\phi$ denotes the MLP parameterized with $\phi$. We generate adjacency matrix by controlling synthetic feature through $g_\phi$. Then we use hyper-parameter $\rho$ to control the sparsity of the adjacency matrix.

\subsection{Roust Learning via Shock Absorber}

\textbf{Min-Max (Adversarial) Optimization.} Starting from the conclusion of gradient matching at time point ${t^{-}}$, we introduce our shock absorber operator to enhance the gradient matching process and thereby expand the optimization space for exploration at time point $t$. We propose to regularly and automatically learn to add a perturbation $\delta$ (generated by adversarial training and refer to shock absorber) on attributes of synthetic graph ${{\cal S}_t}$. Further, we update our adversarial training framework via the following min-max optimization framework:


\vspace{-0.5em}
\begin{equation} 
\footnotesize
\label{adv} 
\begin{aligned}
& \mathop{\min}\limits_{{\theta_{t + 1}}} \mathbb{E}_{\theta_{0} \sim P_{\theta_{0}}}\left\{ \mathop{\max}\limits_{\theta_t^*, \left\| \delta \right\|_p \le \varepsilon} D \Bigg( \nabla_{\theta_{t + 1}}  \ell_t^{\mathcal{S}}, \;\; \nabla_{\theta_{t + 1}} \ell_t^{\mathcal{T}} \Bigg) \right\} \\
& \ell_t^{\mathcal{S}}:=\ell_t^{\mathcal{S}} \Big( f_{\theta_t^*}(\mathcal{S}_t + \delta_\gamma) \Big) \;\;\;\; \ell_t^{\mathcal{T}}:=\ell_t^{\mathcal{T}} \Big( f_{\theta_t^*}(\mathcal{T}), \rm{Y} \Big)
\end{aligned}
\end{equation}
\vspace{-0.5em}

\noindent where $f_{\theta _t^* }$ represents the GNN model parameterized with the fixed optimal ${\theta^*}$ at the $t$-th iteration, $ {\left\|  \cdot  \right\|_p} $  is some $ {\ell _p} $-norm distance metric, $ \varepsilon  $ is the perturbation budget, and $D$ is the distance function which comes from Eq.~\ref{distance}. ${\mathbb{E}_{\theta_{0} \sim P_{\theta_{0}}}}$ denotes that multiple times initialization (satisfies $P_{\theta_{0}}$ distribution) and calculating expectation. Thoroughly achieving Eq.~\ref{adv} need to find a \textbf{temporary and intrusive} variable, \textit{e.g.,} Shock Absorber, which can help to explore the gradients field of the synthetic datasets as much as possible with limited scope. Towards this end, we resort to previous research \cite{madry2017towards}, which has demonstrated that the saddle-point optimization problem of Eq \ref{adv} can be effectively solved using Stochastic Gradient Descent (SGD) for the outer minimization and Projected Gradient Descent (PGD) for the inner maximization. Similarly, the approximation of the inner maximization under an $l_\infty$ constraints is as follows:


\vspace{-0.7em}
\begin{equation} \footnotesize
\label{shockab}
{{\bf{\delta }}_{\gamma  + 1}} = {\Pi _{{\bf{||\delta|| }}{_\infty } \le {\rm{\varepsilon}}}}\left( {{{\bf{\delta }}_\gamma } + \alpha \cdot D\left( {{\nabla _{{\theta ^* _t}}}\ell _t^{\rm{{\cal S}}},{\nabla _{{\theta ^* _t}}}\ell _t^{\rm{{\cal T}}}} \right)} \right) \;\; 
\end{equation} 
\vspace{-0.7em}



\noindent where the perturbation ${\bf{\delta }}$ is updated iteratively for $M$ round, and the function ${\Pi _{{\bf{\delta }}{_\infty } \le {\varepsilon}}}$ performs projection onto the ${\varepsilon}$-ball in the ${l_\infty }$-norm. Compared to the traditional adversarial attack or training algorithms \cite{kong2022robust, madry2017towards}, We removed the sign function, as we aim for updates to be within a more granular range, the diversified perturbations can ensure the process is more robust.

\begin{figure}
 \centering
 \includegraphics[width=1.0\columnwidth]{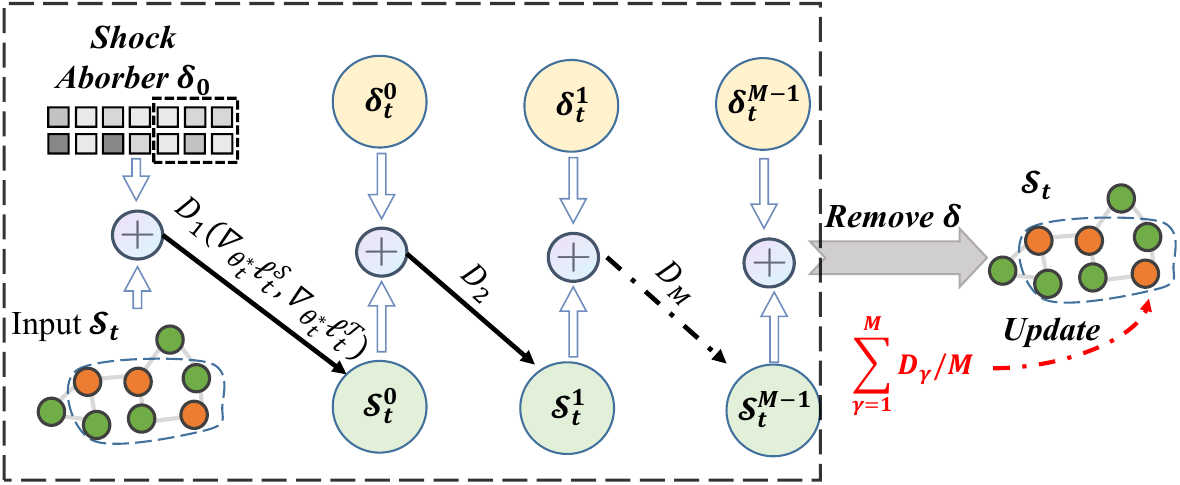}
 \vspace{-0.5em}
 \caption{An illustration of working process of Shock Absorber}
 \vspace{-1.0em}
 \label{fig:SA}
\end{figure}
 

We iteratively update $M$ times to generate the perturbations as depicted in Eq.~\ref{shockab}, this process requires $M$ end-to-end forward and backward passes. For ease of understanding, we illustrate the process through which the shock absorber operates in Fig.~\ref{fig:SA}. In $M$ rounds updating, we iteratively fuse the perturbations with synthetic graph. 

In this fashion, the most severe perturbations ${{\bf{\delta }}_M}$ are applied to the input features, upon which the model weights are optimized. It is worth emphasizing that we have no parameter update process in this procedure, we preserve the parameter gradient throughout the entire M iterations and subsequently eliminate this perturbation after M rounds of Shock Absorber influence. In next time point ($t^{+}$), we use the following function to proceed to update the synthetic dataset:

\vspace{-0.5em}
\begin{equation} \scriptsize
    D_{{t^ + }}^{\rm{{\cal S}}} = \frac{1}{M}\sum\limits_{\gamma  = 1}^M {D_t^\gamma \left( {{\nabla _{\theta _t^*}}\ell _t^{\rm{{\cal S}}},{\nabla _{\theta _t^*}}\ell _t^{\rm{{\cal T}}}} \right)} 
\end{equation}
\vspace{-0.5em} 

\noindent Here $D_t^\gamma $ denotes distance at $\gamma$ round at $t$ points. $D_{{t^ + }}^{\rm{{\cal S}}}$ represents the distance after attaching $M$ round shock absorber. At $t^{+}$ time point, we only use the average gradient values to update the synthetic datasets.

\vspace{-0.4em}
\subsection{A time-efficient version of GroC}
To better generalize to large datasets and reduce the computational complexity, we provide a time-efficient version of GroC called TimGroC. Compared to GroC, TimGroC achieves a significantly efficient time benefit, enhancing model robustness with almost no additional time overhead. In the implementation of TimGroC, we removed the training loop that updates the adversarial perturbation during adversarial training optimization (\textit{i.e.}, as illustrated in Fig \ref{fig:SA}). This allows the $M$ iterations at time $t^ +$ to be integrated into the outer loop for optimizing the synthesized dataset. Specifically, the adversarial perturbation is set as a persistent variable and added to the synthesized data for gradient matching. This process involves both forward and backward passes, simultaneously obtaining gradients for the synthesized dataset and the shock absorber. This process can be understood as free adversarial training \cite{shafahi2019adversarial}. Based on this, we reap the benefits brought by adversarial training, with virtually no additional time cost introduced.



%


\vspace{-0.5em}
\subsection{Gradient Locating in Synthesized Data}
\vspace{-0.5em}

In this work, we focus on using the shock absorber for helping the graph data condensation process be more robust. However, employing excessively large perturbations may diminish the expressive power of the entire synthetic dataset. Therefore, \textbf{we selectively apply perturbations solely to the most vulnerable portion of the synthetic dataset.} We refer to this process as gradient localization, as it involves identifying the optimal location for applying perturbations. Concretely, we perform element-wise multiplication between a differentiable all-ones matrix $m_\delta$ and the perturbations $\delta$. The purpose of operation is to incorporate the all-ones matrix into the optimization computation. Note that the shape of $\delta$ is identical to that of the synthetic data $\mathcal{S}$.


 \vspace{-0.3em}
\begin{equation} \scriptsize
R = \left| {{\nabla _{{m_\delta }}}\left( {D\left( {{\nabla _\theta }\ell ({f_{{\theta _t}}}({\rm{{\cal S}}} + \delta  \odot {m_\delta } \odot {m_g}),{{\rm{Y}}^\prime }),{\nabla _\theta }\ell ({f_{{\theta _t}}}({\rm{{\cal T}}}),{\rm{Y}})} \right)} \right)} \right|
\end{equation} 
 \vspace{-0.5em}

\noindent where $R$ is the absolute value of the gradient of the ${m_\delta }$. The adversarial training is performed $M$ times backward pass and forward pass, the noise $\delta _0$ is uniform noise, ${m_g}$ is given an initial value of all one matrix. Through the above gradient information, we use topk algorithm to get position ${m_g}$:
 \vspace{-0.5em}
\begin{equation}
{m_g}_{(i,j)}= \begin{cases}1, & \text { if } R_{i, j} \text { are the top-$k$ largest entries} \\ 0, & \text { otherwise }\end{cases}
\end{equation}
 \vspace{-0.5em}

\noindent The ${m_g}$ obtained by the gradient in the previous round acts on the noise ${\delta _{\gamma + 1}}$ of the second round

 \vspace{-0.5em}
\begin{equation}
{\delta _{\gamma}} = {\delta _\gamma ^\prime} \odot {m_{\delta}} \odot {m_{g,\gamma}}
\end{equation}
 \vspace{-0.5em}

\noindent where the ${m_{g,0}}$ is the all-ones matrix, after computing the gradient information for the position matrix ${m_\delta }$ at the first step, it becomes a sparse matrix with only 0 and 1 entries. Therefore, after incorporating our method into Eq.~\ref{eq.inner_para}, it can be rewritten as:

 \vspace{-0.5em}
\begin{equation} \footnotesize
\begin{aligned}
    & \mathop {\min }\limits_{\rm{{\cal S}}} \sum\limits_{t = 0}^{T - 1} {D\left( {{\nabla _\theta }\ell ({f_{{\theta _t}}}({\rm{{\cal S}}} + \delta  \odot {m_\delta } \odot {m_g}),{{\rm{Y}}^\prime }),{\nabla _\theta }\ell ({f_{{\theta _t}}}({\rm{{\cal T}}}),{\rm{Y}})} \right)} 
\end{aligned}
\end{equation} 
 \vspace{-0.5em}

\vspace{-0.4em}
\section{Experiments}
\vspace{-0.2em}

We present empirical results to demonstrate the effectiveness of our proposed methods GroC and TimGroC. The experiments aim to answer the following research questions:

\begin{itemize}[leftmargin=*]
    \item \textbf{RQ1.} How is the evaluation quality of GroC and TimGroC compared to that of existing SOTAs?
    
    \item \textbf{RQ2.} How effective is shock absorber?

    \item \textbf{RQ3.} What is the time overhead of our model?
    
    \item \textbf{RQ4.} Does the graph compressed by our model exhibit transferability across backbones?
    
\end{itemize}

\vspace{-0.4em}
\subsection{Experimental Setting}

\begin{table*}[htbp]
	\centering
 \caption{Comparison between condensed graphs and original graphs. The condensed graphs have
fewer nodes and are more dense.}
\vspace{-0.5em}
\resizebox{1\linewidth}{!}{
\begin{tabular}{lcccccccccccc}
\toprule
  \multirow{2}{*}{}&\multicolumn{3}{c}{\textbf{Citeseer}, $r$=0.3\%} 
  &\multicolumn{3}{c}{\textbf{Cora}, $r$=0.4\%}
  &\multicolumn{3}{c}{\textbf{ogbn-arxiv}, $r$=0.05\%}
  &\multicolumn{3}{c}{\textbf{Flickr}, $r$=0.1\%}
  \\
\cmidrule(lr){2-4} \cmidrule(lr){5-7} \cmidrule(lr){8-10} \cmidrule(l){11-13}
  & \textbf{GroC/TimGroC} & \textbf{GCond} & \textbf{Original} & \textbf{GroC/TimGroC} & \textbf{GCond} & \textbf{Original} & \textbf{GroC/TimGroC} & \textbf{GCond} & \textbf{Original} & \textbf{GroC/TimGroC} & \textbf{GCond} & \textbf{Original} \\
  \midrule
 Accuracy (\%) & 67.82/69.16  & 64.13  & 71.12   & 80.91/82.02  & 80.61   & 80.91  & 58.46/57.62  & 56.49  & 70.76  & 46.95/47.01   & 46.93  & 47.16     \\

  Nodes & 6 & 6  & 3327  & 7  & 7  & 2708  & 90  & 90  & 169343  & 44  & 44  & 44625  
\\
  Edges & 15 & 15  & 4732  & 21  & 21  & 5429  & 3880  & 3955  & 1166243  & 946  & 946  & 218140  
\\
 Sparsity & 83.33 \% & 83.33 \%  & 0.09 \%  & 85.71 \%  & 85.71 \%  & 0.15 \%  & 95.80 \%  & 97.65 \%  & 0.01 \%  & 97.73 \%  & 97.73 \%  & 0.02 \%  
\\
  Storage & 0.085 MB & 0.087 MB  & 47.1 MB  & 0.041 MB  & 0.041 MB  & 14.9 MB  & 0.078 MB  & 0.078 MB  & 100.4 MB  & 0.094 MB  & 0.094 MB  & 86.8 MB  
  \\\hline
\end{tabular}
}
 \label{main1}
 \vspace{-0.5em}
\end{table*}

\begin{figure*}[h]
	\begin{center}
		\begin{tabular}{ccccc}
			\includegraphics[width=0.37\columnwidth, frame]{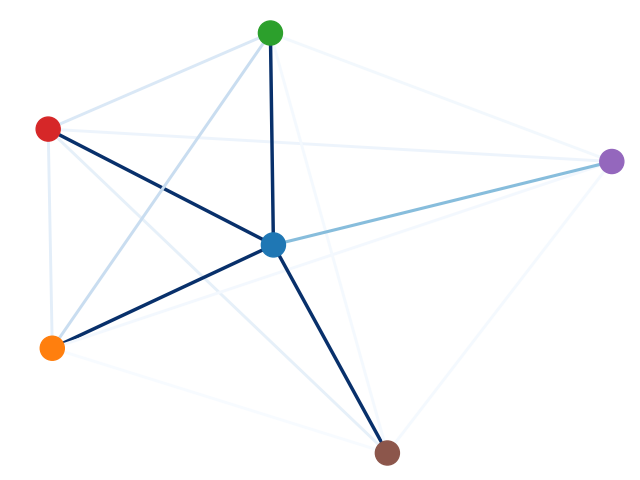}&
			\includegraphics[width=0.37\columnwidth, frame]{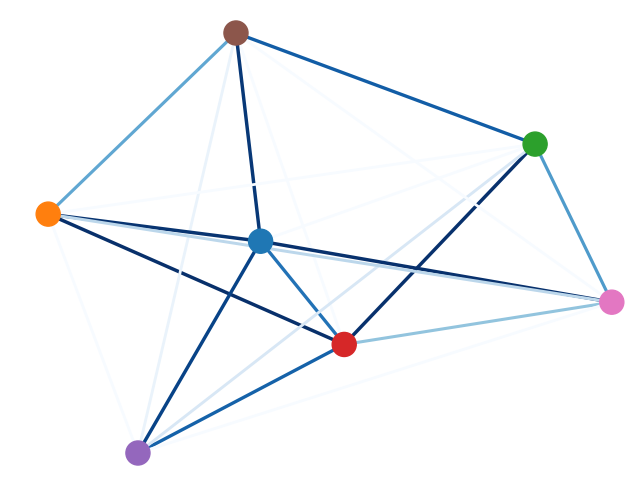}&
            \includegraphics[width=0.37\columnwidth, frame]{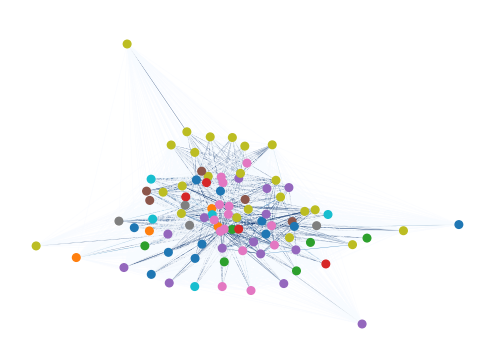}&
            \includegraphics[width=0.37\columnwidth, frame]{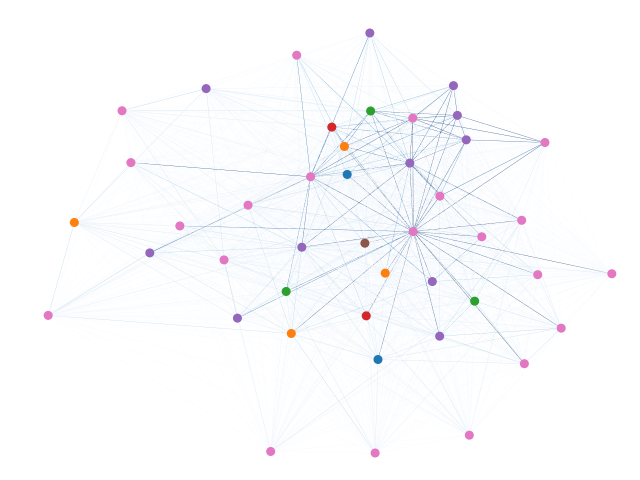}&
            \includegraphics[width=0.37\columnwidth, frame]{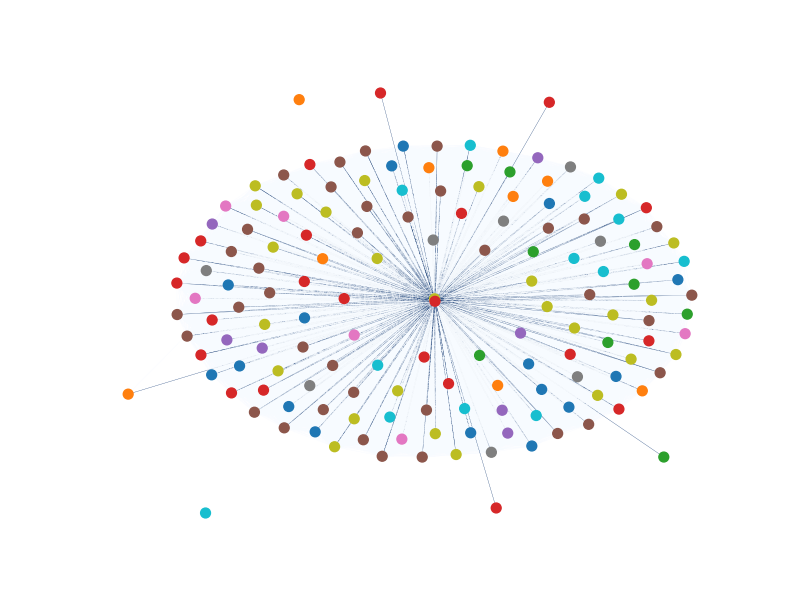}\\
			{\scriptsize (a) Citeseer, $r$=0.3\%}& {\scriptsize (b) Cora, $r$=0.4\%} & {\scriptsize (c) Ogbn-arxiv, $r$=0.05\%}& {\scriptsize (d) Flickr, $r$=0.1\%}& {\scriptsize (d) Reddit, $r$=0.1\%} \\
		\end{tabular}
	\end{center}
	\vspace{-0.4cm}
	\caption{\textbf{Visualization of condensed graphs.} The learned condensed graphs are weighted graphs, The color depth of the edges varies with the weight.}
	\label{fig:adj_vis}
\end{figure*}

\textbf{Datasets \& Baselines.} We conduct experiments on three transductive datasets, i.e., Citeseer, Cora \cite{kipf2016semi} and Ogbn-arxiv \cite{hu2020open} and on the inductive dataset, i.e., Flickr \cite{zeng2019graphsaint} and Reddit \cite{hamilton2017inductive}. For the datasets setting, we follow the setup in \cite{jin2021graph}. In addition, we also examine the transfer ability of our Shock Absorber on the graph classification task. We utilize the Ogbg-molhiv molecular dataset from Open Graph Benchmark (OGB) \cite{hu2020open} and TUDatasets (DD and NCI1) \cite{DBLP:journals/corr/abs-2007-08663} for graph-level property classification. On node classification datasets, We compare our method with one state-of-the-art condensation method and three coreset methods. (\romannumeral1) GCond \cite{jin2021graph} models the condensed graph structure based on the condensed node features. (\romannumeral2) Random coreset \cite{welling2009herding} randomly selects nodes for graph sampling. (\romannumeral3) The Herding coreset \cite{welling2009herding} is often used in continual learning to select samples closest to the cluster center.(\romannumeral4) The K-Center method \cite{farahani2009facility,DBLP:conf/iclr/SenerS18} minimizes the maximum distance between a sample and its nearest center to select center samples. For the four baselines: Random, Herding, K-Center, and GCond, we use the implementations from \cite{jin2021graph}.

\textbf{Evaluation.} We train our method and SOTAs with the same settings, including learning rate, optimizer, etc. Firstly, we create three condensed graphs by training methods with different seeds. Then, we train a GNN on each graph, repeating the process three times. To assess condensed graph information, we train GNN classifiers and evaluate on real graph test nodes or graphs. By comparing model performance on real graphs, we obtain condensed graph informativeness and effectiveness. Experiments are repeated 3 times, reporting average performance and variance.


\textbf{Backbones.} To ensure fairness, we utilize the identical model as GCond \cite{jin2021graph}, specifically GCN \cite{kipf2016semi}, for evaluation. In the condensation process, we apply SGC \cite{wu2019simplifying}, configuring it as a 2-layer model with 256 hidden units.


\subsection{Main Results (RQ1)}

In this part, we thoroughly investigate the performance of the GroC and TimGroC across various datasets. We conduct a comprehensive comparison of our frameworks with Random, Herding, K-Center, and GCond, for node classification tasks on the Cora, Citeseer, Ogbn-arxiv and Flickr datasets. In Tab~\ref{main1}, we present a comparison of sparsity performance and other related parameters between our model and the current state-of-the-art models. In Tab~\ref{main2}, we further extend the comparison by including several clustering/compression algorithms. The observations (\textbf{obs}) can be listed as follows.

\begin{figure*}[htbp]
	\begin{center}
		\begin{tabular}{ccccc}
			\includegraphics[width=0.35\columnwidth]{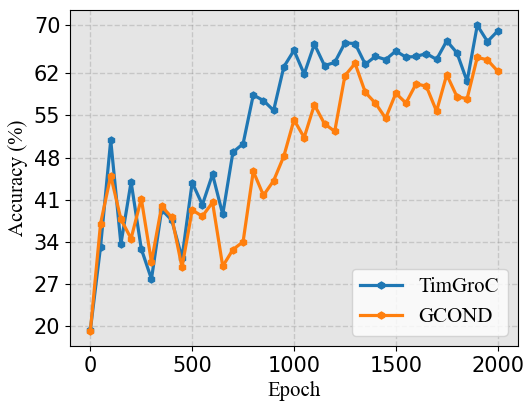}&
			\includegraphics[width=0.35\columnwidth]{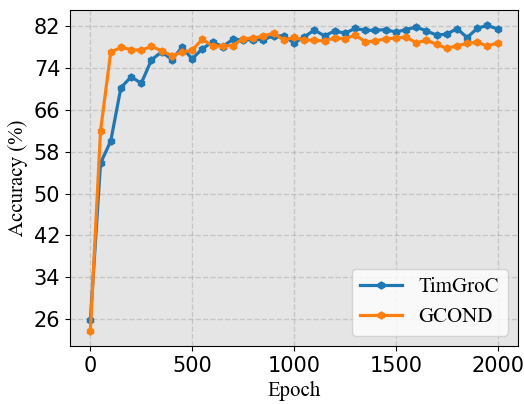}&
            \includegraphics[width=0.35\columnwidth]{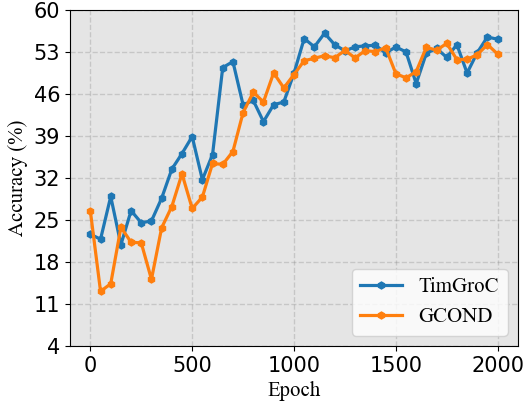}&
			\includegraphics[width=0.35\columnwidth]{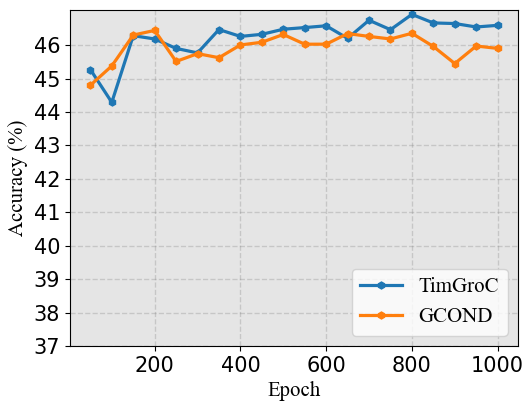}&
			\includegraphics[width=0.35\columnwidth]{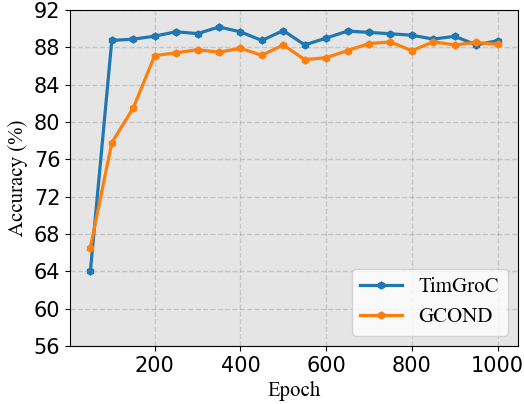} \\
			{\scriptsize (a) Citeseer, $r$=0.3\%}& {\scriptsize (b) Cora, $r$=0.4\%} & {\scriptsize (c) Ogbn-arxiv, $r$=0.05\%} & {\scriptsize (d) Flickr, $r$=0.1\%} & {\scriptsize (d) Reddit, $r$=0.1\%} \\
		\end{tabular}
	\end{center}
	\vspace{-0.4cm}
	\caption{\textbf{Performance during training.} The test performances (accuracy) on Citeseer, Cora, Ogbn-arxiv and Flickr, Reddit.}
	\label{fig:feature_vis}
\end{figure*}

\begin{table*}[htbp]
  \caption{Shock Absorber without gradient location (GL) and Shock Absorber achieves promising performance in comparison to baselines even with extremely large condensation rates. We choose GCN as backbones and report the mean and standard deviation of the results of transductive performance on Citeseer, Cora, Ogbn-arxiv. Performance is reported as test accuracy (\%).  }
  \vspace{-1em}
\setlength{\tabcolsep}{2.5pt}
\begin{center}
\small
{
 \begin{tabular}{cccccccccc} 
 \toprule
    \multirow{2}{*}{\bf Dataset}  & \multirow{2}{*}{\tabincell{c}{\bf Ratio}} &  \multicolumn{4}{c}{\bf Baselines} &  \multicolumn{1}{c}{\bf Ablation} &  \multicolumn{3}{c}{\bf Ours } \\ 
    \cmidrule(lr){3-7} \cmidrule(l){8-10}
    
     &  & \bf Random & \bf Herding & \bf K-Center & \bf GCond  & \bf w/o GL & \bf GroC  & \bf TimGroC  & \bf Full graph \\ 
     
     \midrule
     
       
    

    
     Citeseer & $0.3\% $ & ${33.87_{\pm 0.82}}$ &${31.31_{\pm 1.20}}$ & ${34.03_{\pm 2.52}}$ &${64.13_{\pm 1.83}}$ & ${63.98_{\pm 4.31}}$ & \cellcolor{gray!30}${67.82_{\pm  1.31}}$ & \cellcolor{gray!30}${69.16_{\pm  2.00}}$ & ${71.12_{\pm  0.06}}$\\
     
     Cora & $0.4\% $ & ${37.04_{\pm 7.41}}$ & ${43.47_{\pm 0.55}}$ & ${46.33_{\pm 3.24}}$ & ${80.61_{\pm 0.31}}$ & ${80.43_{\pm 0.45}}$ & \cellcolor{gray!30}${80.91_{\pm 0.39}}$ & \cellcolor{gray!30}${82.02_{\pm 0.42}}$ & ${80.91_{\pm  0.10}}$ \\
     
     Ogbn-arxiv & {$0.05\% $} & {$46.83_{\pm 2.60}$} & {${49.74_{\pm 2.30}}$} & {${47.28_{\pm 1.15}}$} & {${56.49_{\pm 1.69}}$} & {${57.39_{\pm 0.65}}$} & \cellcolor{gray!30}{${58.46_{\pm 0.85}}$}  & \cellcolor{gray!30}{${57.62_{\pm 0.64}}$}  & {${70.76_{\pm  0.04}}$}  \\

     Flickr & {$0.1\% $} & {$41.84_{\pm 1.87}$} & {${43.90_{\pm 0.56}}$} & {${43.30_{\pm 0.90}}$} & {${46.93_{\pm 0.10}}$} & {${46.81_{\pm 0.10}}$} & \cellcolor{gray!30}{${46.95_{\pm 0.03}}$}  & \cellcolor{gray!30}{${47.01_{\pm 0.10}}$}  & {${47.16_{\pm  0.17}}$}  \\

     Reddit & {$0.1\% $} & {$59.14_{\pm 2.26}$} & {${65.75_{\pm 1.28}}$} & {${53.05_{\pm 2.73}}$} & {${89.34_{\pm 0.54}}$} & {${89.56_{\pm 0.74}}$} & \cellcolor{gray!30}{${89.56_{\pm 0.45}}$}  & \cellcolor{gray!30}{${90.24_{\pm 0.05}}$}  & {${93.96_{\pm  0.03}}$}  \\
  \midrule
  \end{tabular}
  }
  \vspace{-1pt}

  \label{main2}
   \end{center}
   \vspace{-6mm}
\end{table*}

\begin{table}[!h] 
\footnotesize
\setlength{\tabcolsep}{1.8pt}
  \caption{The shock absorber, and without gradient location (GL) compared to DosCond. We report the accuracy for the first two datasets and ROC-AUC for the last dataset. We record the original graph performance of three graph as 78.81, 70.88 and 74.17.} 
  \vspace{-0.7em}
  \centering
  \begin{tabular}{ccccccc}
    \toprule
     \textbf{Dataset}   &  \textbf{Ratio}  &  \textbf{DoSCond}&   \textbf{w/o GL} &  \textbf{Shock Absorber}   \\
    \midrule
    DD   & $0.2\%$ & ${72.31_{\pm 3.81}}$ & ${73.73_{\pm 1.06}}$ & \cellcolor{gray!30}${74.29_{\pm  1.06}}$   \\
    
      NCI1  & $0.1\%$  & ${57.58_{\pm  0.13}}$ & ${57.93_{\pm 0.14}}$ & \cellcolor{gray!30}${58.20_{\pm 0.39}}$    \\
      Ogbg-molhiv  & $0.01\%$   & ${73.81_{\pm  0.11}}$ & ${74.08_{\pm 0.34}}$ &\cellcolor{gray!30} ${ 74.17_{\pm 0.10}}$ &   \\

    \bottomrule
  \end{tabular}\label{tab:at2}
  \vspace{-1em}
\end{table}

\vspace{-0.2em}
\begin{itemize}[leftmargin=*]
    \item \textbf{Obs 1:} \textbf{GroC/TimGroC consistently outperform GCond} under extremely large condensation rates, verifying its extraordinary performance. For instance, on the Citeseer dataset, our model can achieve a compression rate under 0.003, which is nearly 5.2\% higher than the current SOTA GCond. These results demonstrate the significance of adversarial training to graph condensation (Tab.~\ref{main1}). Interestingly, from our visualization results (Fig \ref{fig:adj_vis}) and Table \ref{main1}, it can be observed that the graphs we compressed are highly dense, with edges acting as dense information carriers, which facilitates information storage.

\vspace{-0.3em}
    \item \textbf{Obs 2:} \textbf{GroC demonstrates superior performance and lower variance, which attests to the robustness of our algorithm.} For instance, in Tab \ref{main2}, both GroC and TimGroC achieved nearly the lowest variance and the highest performance. On the Citeseer dataset, they show an improvement of almost 5.2\% accompanied by a decline in variance of nearly 2.0\%. Meanwhile, on the Reddit, our framework exhibits a variance of only 0.05\%, which is significantly lower than other models, further corroborating the robustness of our GroC and TimGroC frameworks.

\vspace{-0.2em}
    \item \textbf{Obs 3:} \textbf{GroC exhibits stronger training robustness.} GroC exhibits stronger training robustness. In Fig \ref{fig:feature_vis}, it is readily apparent that, during the training phase, GroC's curve is predominantly above that of GCond, demonstrating significant potential. Particularly on the Citeseer and Ogbn-Arixv datasets, our model excels and surpasses the current best models, still possessing the capability for further improvement towards the end of the training.

\end{itemize}

\subsection{Scability of Shock Absorber (RQ2)}

Additionally, we evaluate Shock Absorber and DoSCond on the graph classification task using the Ogbg-molhiv \cite{hu2020open}, DD, and NCII datasets \cite{morris2020tudataset}. This comparative analysis allows us to assess the effectiveness and superiority of the shock absorber method in different scenarios. Following the methodology of DoSCond \cite{jin2022condensing}, a condensation method for graph classification, we generated one condensed graph for each class. In the training phase of condensed graphs, we integrated our proposed shock absorber into the one-step gradient matching process. For this purpose, we employed a 3-layer GCN for gradient matching. During the test phase, we used a GCN with the same architecture and trained the model on condensed graphs for 500 epochs with learning rate 0.001. The results of the classification accuracy are summarized in Tab.~\ref{tab:at2}.


\noindent \textbf{Scability in Graph Classification.} As shown in Tab. \ref{tab:at2}, in the graph classification scenario, our GroC shows a decent generalization. Its effectiveness can be attributed to the adversarial perturbation which improves the robustness during the graph condensation process. With the gradient localization in synthesized data, condensed graphs contain more effective information which is beneficial for model training. Moreover, our experiments on graph-level property classification have demonstrated superior interpretability and generalization ability for graph classification, surpassing leading baselines.

\vspace{-0.5em}
\subsection{Study of time consumption (RQ3)}
\vspace{-0.5em}
In this subsection, we examine the time cost of our algorithm through experiments to further assess whether our model introduces excessive time overhead while enhancing robustness (3070 GPU). Since our aim is to enhance robustness, our model incorporates an adversarial training process. We observed that the model achieves optimal performance when $M$ is between 3 and 4. Consequently, we adopted the more efficient $M=3$ as the setting for GroC to compare time efficiency. We discovered that TimGroC, compared to GroC, achieves an improvement ranging from $3.19 \sim 4.11$ times while maintaining optimal performance. This further substantiates that our algorithm enhances robustness without introducing additional computational power.

\begin{table}[h] 
\footnotesize
\setlength{\tabcolsep}{1.0pt}
  \caption{Comparing the time consumption of (Tim)GroC and GCond. All results are in seconds should be multiplied by 100.} 
  \vspace{-0.7em}
  \centering
  \begin{tabular}{ccccccc}
    \toprule
     \textbf{Dataset}   &  \textbf{Ratio}  &  \textbf{GCond}&   \textbf{TimGroC w/o GL} &  \textbf{GroC}  &  \textbf{TimGroC}  \\
    \midrule
    Cora   & $0.4\%$ & 29.96 & 30.04 & 100.44 & 30.09  \\
    
    Citeseer  & $0.3\%$  & 29.77 & 30.11  & 102.37  & 30.40    \\
      
   Ogbn-arxiv & $0.05\%$   & 182.34 & 183.20 & 582.37 & 182.47 &   \\

    Flicker  & $0.1\%$   & 8.50 & 8.51 & 34.26 & 8.54   \\

    Reddit  & $0.1\%$   & 51.44 & 52.01 & 214.82 & 52.27   \\
    \bottomrule
  \end{tabular}\label{tab:time}
  \vspace{-1em}
\end{table}

\vspace{-0.5em}
\subsection{Study of transferability (RQ4)}
\vspace{-0.2em}
Lastly, we employed GCN as the training backbone and trained the synthesized smaller graph on this new backbone to evaluate its transferability. As shown in Tab \ref{tab:trans}, we choose Cora and Citeseer as benchmarks and follow the reduction ratio of \cite{jin2021graph}, we can easily observe that our synthesized data also achieves commendable performance on GraphSAGE, SGC, and MLP. This further attests to the excellent transferability of our compression algorithm, offering a reliable solution for future data compression.

\begin{table}[h] 
\footnotesize
\setlength{\tabcolsep}{1.8pt}
  \caption{Transferability of our framework, here we choose TimGroC as compressed algorithm.} 
  \vspace{-0.7em}
  \centering
  \scalebox{0.70}{
  \begin{tabular}{cccccccc}
    \toprule
     \multirow{2}{*}{\textbf{Method}}   &   \multirow{2}{*}{\textbf{Ratio}}   &  \multicolumn{3}{c}{\textbf{Baseline}}  &  \multicolumn{3}{c}{\textbf{TimGroC}} \\
     \cmidrule(lr){3-5} \cmidrule{6-8}
     & & \textbf{GraphSAGE} &  \textbf{SGC}  &  \textbf{MLP} & \textbf{GraphSAGE} &  \textbf{SGC}  &  \textbf{MLP} \\
    \midrule
    Cora   & $2.6\%$  & ${80.16_{\pm  0.98}}$ & ${77.36_{\pm  1.13}}$ & ${74.16_{\pm  6.07}}$  & ${79.24_{\pm  1.74}}$ &${78.15_{\pm  0.73}}$ & ${75.07_{\pm  5.29}}$ \\
    
    Citeseer  & $1.8\%$    & ${71.10_{\pm  0.61}}$  & ${66.87_{\pm  1.32}}$ & ${59.32_{\pm  4.15}}$ & ${72.21{\pm  0.35}}$ & ${71.51_{\pm  0.68}}$ & ${65.31_{\pm  3.59}}$ \\
      


    \bottomrule
  \end{tabular}\label{tab:trans}
  }
  \vspace{-1em}
\end{table}

\section{Conclusion}
  \vspace{-0.3em}
In this study, we introduce GroC, a robust adversarial training-based graph condensation framework. GroC leverages principled adversarial training (min-max optimization) to explore the parameter space surrounding the influential parameters in the original matching process. Based on this, we further introduce the shock absorber operator, which enhances the gradient matching process and maximizes the exploration of synthetic dataset gradients within a limited scope. Our experimental results demonstrate that our approach surpasses other SOTA methods in terms of accuracy and efficiency across multiple node classification datasets and graph classification datasets. The evaluation highlights that our condensed graphs effectively retain important structural properties of the original graphs while significantly reducing dimensionality and computational complexity. 




{
\bibliography{reference}
}



\clearpage

\appendix
\onecolumn
\section{Notations} \label{notations}
\vspace{-0.5em}
\begin{table}[htp]\footnotesize
  \centering
  \caption{The notations commonly reported in this work are placed here.}
   \setlength{\tabcolsep}{24pt} 
  \resizebox{0.8\linewidth}{!}{
    \begin{tabular}{cc}
    \toprule
    Notation & Definition \\
    \midrule
    $C$                 & The number of classes of the raw graph \\
    $K$                 & The initialization times  \\
    $M$                 & The number of adversarial perturbation optimizations \\
    $T$     & The training epochs \\
    ${\rm{{\cal T}}}=(\rm{A}, \rm{X}, \rm{ Y})$   & Original graph  \\
    $\rm{{\cal S}}=(\rm{A}^\prime,\rm{X}^\prime,{\rm{Y}^\prime})$     & Synthetic graph  \\
    $D(\cdot)$                 & The distance function for computing gradient loss  \\
    $\rho$   & The hyperparameter to control the  sparsity of adjacency matrix. \\

    \bottomrule
    \end{tabular}%
    }
  \label{tab:addlabel}%
\end{table}%

\section{Algorithm of our GroC method} \label{algo}
\vspace{-0.5em}
\begin{algorithm}[H]
\caption{The framework of our proposed GroC.}
\label{alg:GROc}
\begin{algorithmic}[1]
\STATE \textbf{Input:} Training data Graph ${\rm{{\cal T}}}=(\rm{A}, \rm{X}, \rm{Y})$, pre-defined labels ${\rm{Y}^\prime }$ for condensed synthetic graph. Hyperparameters for control  the progress of synthesized data ${{\rm{{\omega}}}_1}$ and ${{\rm{{\omega}}}_2}$. Learning rates $\eta _1$, $\eta _2$, $\alpha$. $\rho$ is the hyperparameter to control the sparsity of the synthesized data adjacency matrix.

\STATE Randomly selecting node features of raw data to construct synthetic node attributes (one class chooses one node) $\rm{X}^\prime$.

\STATE Initialize ${\delta _0}$ as uniform noise
\STATE Initialize ${m_\delta }$ as a differentiable matrix of all ones
\STATE Initialize ${m_g}$ as a matrix of all ones
\STATE Initialize $\gamma = 0$
\FOR{$k = 0,...,K - 1$}
    \STATE Initialize ${\theta _0} \sim {P_{{\theta _0}}}$
    \FOR{$t = 0,...,T - 1$}
    \STATE ${D^\prime } = 0$
    \IF{$\gamma \;\% M ==0$}
        \STATE Initialize ${\delta _0}$ as uniform noise
        \STATE Initialize ${m_\delta }$ as a differentiable matrix of all ones
        \STATE Initialize ${m_g}$ as a matrix of all ones
        \STATE Initialize $\gamma = 0$
    \ENDIF
    \STATE $\gamma = \gamma+1$
        \FOR{$c = 0,...,C - 1$}
        
            \STATE Compute ${\rm{A^\prime} } = {g_\phi }\left( {\rm{{X^\prime }}} \right)$\; then $\rm{{\cal S}} = (\rm{A}^\prime, \rm{X}^\prime, {\rm{{Y}}^\prime })$ 
            
            \STATE Sample $\left( {{\rm{A}_c},{\rm{X}_c},{\rm{Y}_c}} \right) \sim {\rm{{\cal T}}}$ and $\left( {{\rm{A_c}^\prime },{\rm{X_c}^\prime },{\rm{Y_c}^\prime }} \right) \sim {\rm{{\cal S}}}$

            \STATE Compute ${\ell_t ^{\rm{{\cal T}}}} = \ell ({f_{{\theta _t}}}({\rm{A}_c},{X_c}),{\rm{{ Y_c}}})$\;then ${\ell_t ^{\rm{{\cal S}}}} = \ell ({f_{{\theta _t}}}({\rm{A_c}^\prime },{{X_c}^\prime } + {\delta}_\gamma  \odot {m_\delta } \odot {m_g}),{{\rm{{ Y_c}}}^\prime })$

            \STATE ${D^\prime } \leftarrow {D^\prime } + D({\nabla _{{\theta _t}}}{\ell_t ^{\rm{{\cal T}}}},{\nabla _{{\theta _t}}}{\ell_t ^{\rm{{\cal S}}}})$

        \ENDFOR
        \STATE Compute Shock Absorber ${{\bf{\delta }}_{\gamma + 1}} = {{\bf{\delta }}_\gamma} + \alpha \cdot\left( {{\nabla _\delta }D({\nabla _{{\theta _t}}}{\ell_t ^{\rm{{\cal T}}}},{\nabla _{{\theta _t}}}{\ell_t ^{\rm{{\cal S}}}})} \right)$

        \STATE Compute gradient matrix $R = \left| {{\nabla _{{m_\delta }}}D({\nabla _{{\theta _t}}}{\ell_t ^{\rm{{\cal T}}}},{\nabla _{{\theta _t}}}{\ell_t ^{\rm{{\cal S}}}})} \right|$

        \STATE  Set ${m_g}_{\left( {ij} \right)} = 1$ if ${R}_{ij}$ in the highest Top-K(${R}$) values, otherwise 0
        
        \IF{$t\;\% \left( {{{\rm{{\omega}}}_1} + {{\rm{{\omega}}}_2}} \right) < {{\rm{{\omega}}}_1}$}
            \STATE Update ${\rm{X}^\prime } \leftarrow {\rm{X}^\prime } -  \eta _1{\nabla _{{\rm{X}^\prime }}}{D^\prime }$
        \ELSE
            \STATE Update ${g_\phi }  \leftarrow {g_\phi }  - {\eta _2}{\nabla _\phi }{D^\prime }$      
        \ENDIF
        \STATE Update ${\theta _{t + 1}} \leftarrow {\rm{op}}{{\rm{t}}_\theta }\left( {{\theta _t},{\rm{{\cal S}}},{\tau _\theta }} \right)$ 
    \ENDFOR
\ENDFOR

\STATE ${\rm{A}^\prime } = {g_\phi }\left( {{\rm{X}^\prime }} \right)$
\STATE  $\rm{A}_{ij}^{'} = \rm{A}_{ij}^{'}$ if ${{\rm{A}_{ij}}^{'}} > \rho$, otherwise 0
\RETURN Condensed graph $(\rm{A}^\prime,\rm{X}^\prime,{\rm{Y}^\prime})$
\end{algorithmic}
\end{algorithm}

\end{document}